\documentclass[10pt,twocolumn,letterpaper]{article}

\usepackage[pagenumbers]{cvpr}

\newcommand{\tocite}[1]{\textcolor{red}{[TOCITE]}}
\newcommand{\toexp}[1]{\textcolor{blue}{[TOEXP]}}
\newcommand{\PAR}[1]{\vskip2pt \noindent{\bf #1~}}
\newcommand\blfootnote[1]{%
    \begingroup
    \renewcommand\thefootnote{}\footnote{#1}%
    \addtocounter{footnote}{-1}%
    \endgroup
}

\newlength\savewidth

\newcommand{\methodname}{LiDAR-RT}
\newcommand{\methodnameblank}{\methodname\ }

\definecolor{colorfirst}{rgb}{.866,.945, 0.831} %
\definecolor{colorsecond}{rgb}{1, 0.98, 0.83} %
\definecolor{colorthird}{rgb}{0.76, 0.87, 0.92} %

\newcommand{\cellfirst}{\cellcolor{colorfirst}}
\newcommand{\cellsecond}{\cellcolor{colorsecond}}

\usepackage{ragged2e}
\usepackage{booktabs,makecell, multirow, tabularx}
\usepackage{tabularray}
\usepackage{colortbl}
\usepackage{graphicx}
\usepackage{bbding}
\usepackage{pifont}

\usepackage[accsupp]{axessibility}

\definecolor{cvprblue}{rgb}{0.21,0.49,0.74}
\usepackage[pagebackref,breaklinks,colorlinks,citecolor=cvprblue]{hyperref}

\def\confName{CVPR}
\def\confYear{2025}

\title{LiDAR-RT: Gaussian-based Ray Tracing for Dynamic LiDAR Re-simulation}

\author{
    Chenxu Zhou\textsuperscript{1*} \quad
    Lvchang Fu\textsuperscript{2*} \quad
    Sida Peng\textsuperscript{1} \quad
    Yunzhi Yan\textsuperscript{1} \quad
    Zhanhua Zhang\textsuperscript{3} \quad
    \\[2pt]
    Yong Chen\textsuperscript{3} \quad
    Jiazhi Xia\textsuperscript{2} \quad
    Xiaowei Zhou\textsuperscript{1} \quad
    \\[5pt]
    $^1$Zhejiang University \qquad
    $^2$Central South University \qquad
    $^3$Geely Automobile Research Institute
}

\begin{document}

\twocolumn[
    \maketitle
    \vspace{-2em}
    \begin{center}
    \captionsetup{type=figure}
    \vspace{4pt}
    \includegraphics[width=1.0\textwidth]{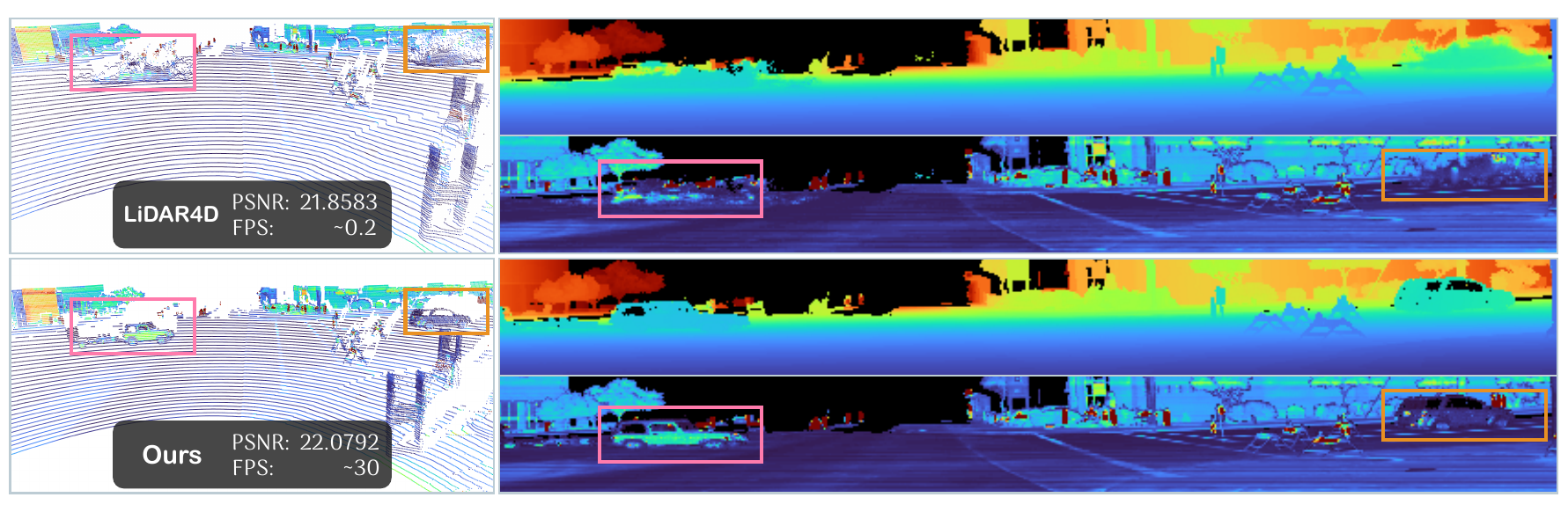}
    \vspace{-6mm}
    \captionof{figure}{%
        \textbf{Realistic and real-time rendering of LiDAR view in dynamic driving scenes.}
        Our \methodnameblank produces high-fidelity LiDAR view at 30 FPS (64×2650) within 2 hours of training.
        SOTA method~\cite{zheng2024lidar4d} struggles to model the dynamic objects in complex scenes and suffers from high training and rendering costs (15 hours for training and 0.2 FPS for rendering a range image).
    }
    \vspace{2pt}
    \label{fig:teaser}
\end{center}

    \bigbreak
]

\blfootnote{$^*$The first two authors contributed equally.}

\tolerance=40
\begin{abstract}
    This paper targets the challenge of real-time LiDAR re-simulation in dynamic driving scenarios.
    Recent approaches utilize neural radiance fields combined with the physical modeling of LiDAR sensors to achieve high-fidelity re-simulation results.
    Unfortunately, these methods face limitations due to high computational demands in large-scale scenes and cannot perform real-time LiDAR rendering.
    To overcome these constraints, we propose \methodname, a novel framework that supports real-time, physically accurate LiDAR re-simulation for driving scenes.
    Our primary contribution is the development of an efficient and effective rendering pipeline, which integrates Gaussian primitives and hardware-accelerated ray tracing technology.
    Specifically, we model the physical properties of LiDAR sensors using Gaussian primitives with learnable parameters and incorporate scene graphs to handle scene dynamics.
    Building upon this scene representation, our framework first constructs a bounding volume hierarchy (BVH),
    then casts rays for each pixel and generates novel LiDAR views through a differentiable rendering algorithm.
    Importantly, our framework supports realistic rendering with flexible scene editing operations and various sensor configurations.
    Extensive experiments across multiple public benchmarks demonstrate that our method outperforms state-of-the-art methods in terms of rendering quality and efficiency.
    Our project page is at \url{https://zju3dv.github.io/lidar-rt}.
\end{abstract}
\vspace{-6mm}

\section{Introduction}
\label{sec:introduction}

Modeling dynamic urban scenes is of great importance for a variety of applications, such as digital twin simulation, autonomous driving, and virtual reality.
Recent research has shown impressive results of novel view synthesis in driving scenarios~\cite{yan2024street, chen2023periodic,Zhou_2024_CVPR, chen2024omnire} with camera sensors, significantly enhancing the fidelity and diversity of data for downstream tasks and applications.
However, the majority of current approaches overlook the re-simulation of LiDAR sensors, which are key components for many 3D perception algorithms, thereby restricting the scope of simulation applications.
In contrast to cameras, the inherent physical properties of LiDAR sensors and sparsity of point clouds make it challenging to reconstruct and simulate, especially in dynamic scenes with complicated motions.

Previous methods~\cite{li2023pcgen,manivasagam2020lidarsim} initially reconstruct 3D scenes from real-world data with explicit representations like dense point clouds or triangular meshes, which are utilized to render novel LiDAR views via ray-casting.
Although these two-stage methods manage to produce acceptable results, they are hampered by geometric inaccuracies and struggle to model the physical properties of LiDAR sensors.
Additionally, they are only capable of modeling static scenes and fail to handle dynamic objects.

As neural scene representations advance, there have been some works~\cite{tao2023lidar, Huang2023nfl, xue2024geonlf, zhang2024nerflidar} that attempt to synthesize novel views of LiDAR with neural radiance fields~\cite{mildenhall2020nerf}.
These NeRF-based methods~\cite{tao2023lidar,Huang2023nfl,xue2024geonlf,zhang2024nerflidar} combine the rendering power of neural fields with the physical modeling process of LiDAR sensors, achieving physically realistic LiDAR rendering.
By introducing 4D hybrid features or tracked bounding boxes, some recent works~\cite{zheng2024lidar4d,Wu2023dynfl} extend these methods to handle dynamic scenes.
However, these approaches are constrained by high training costs and slow rendering speeds due to the large amount of network parameters, and struggle to model complex dynamic scenes with long-distance vehicle motion and occlusions.

In this paper, we propose a novel framework, named \methodname, for novel LiDAR view synthesis of dynamic driving scenes.
Our primary contribution is the development of an efficient and effective rendering pipeline for Gaussian-based~\cite{Kerbl20233dgs, Huang2DGS2024} LiDAR representation.
Firstly, We enhance Gaussian primitives with additional learnable parameters to accurately model the physical characteristics of LiDAR sensors.
Furthermore, to tickle the challenges of dynamic scenes, we integrate the scene graphs~\cite{ost2021nsg} into LiDAR representation, which provides flexible modeling capabilities under various environmental conditions.

Building upon proposed scene representation, we design a differentiable Gaussian-based ray tracer to simulate the physical formation of LiDAR sensors.
Specifically, our method construct the corresponding proxy geometries for Gaussian primitives and insert them into a bounding volume hierarchy (BVH).
To compute the LiDAR radiance, we cast ray for each pixel from the sensor against the BVH to determine the intersections and store the information in a sorted buffer~\cite{Louis2007kbuffer}.
Subsequently, we calculate the response of the intersected Gaussians and integrate the radiance along the ray via volumetric rendering techniques.
This process continues until all Gaussian primitives have been traversed or the accumulated transmittance reaches a predefined threshold.
Then the rendered properties are fused to generate novel LiDAR views.
Compared to the vanilla tile-based Gaussian rasterizer~\cite{Kerbl20233dgs}, our method employs a physically accurate ray tracing process, which enhances both the realism and accuracy for LiDAR re-simulation.
Thanks to the components we proposed above, our method not only reconstructs high fidelity LiDAR point clouds and achieves real-time rendering of novel LiDAR views, but also supporting flexible manipulations of LiDAR sensors.

We evaluate the proposed method on Waymo Open (Waymo)~\cite{sun2020waymo} and KITTI-360~\cite{liao2022kitti} datasets, which cover a wide range of complex dynamic scenes.
Our method achieves state-of-the-art performance in terms of rendering quality and efficiency across all datasets.
Moreover, our comprehensive experiments have confirmed that the components we proposed are highly adaptable for LiDAR re-simulation and maintain great performance under various editing manipulations.

\begin{figure*}[ht]
    \centering
    \vspace{-2mm}
    \includegraphics[width=1.0\linewidth]{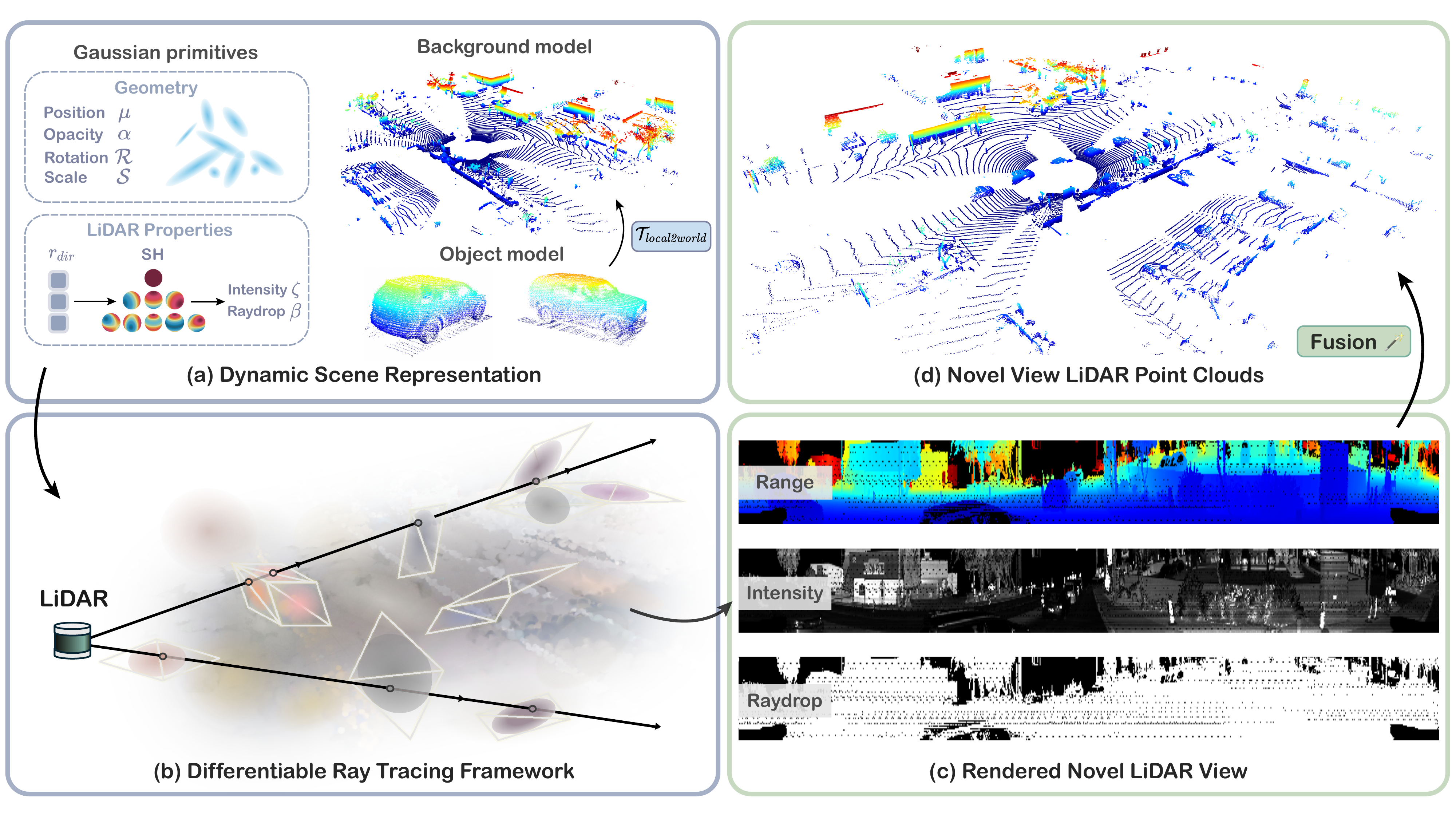}
    \vspace{-8mm}
    \caption{
        \textbf{Overview of \methodname.}
        (a) We decompose the dynamic scene into a background model and multiple object models, with each represented by a set of Gaussian primitives. In addition to geometric attributes, we introduce learnable parameters (SHs) on Gaussians to emulate the intrinsic properties $(\zeta, \beta)$ of LiDAR sensors.
        (b) Based on this representation, we design a differentiable ray tracing framework. We first construct the proxy geometry for Gaussian primitives and then cast rays from the sensor to perform intersection tests.
        (c) By evaluating the response from these intersections, we accumulate point-wise properties along each ray, and finally render the novel LiDAR view as range images.
        (d) The range images are fused and re-projected into LiDAR point clouds for downstream tasks.
    }
    \label{fig:pipeline}
    \vspace{-3mm}
\end{figure*}

\section{Related work}
\label{sec:related_work}

\PAR{LiDAR Simulation.}
Traditional self-driving simulators such as CARLA~\cite{Dosovitskiy17} or Airsim~\cite{shah2017airsimhighfidelityvisualphysical} do possess the ability to simulate LiDAR sensors,
but suffer from the sim-to-real gap~\cite{lindstorm2024sim2realgap,man2023towards} and costly manual effort in creating virtual assets.
LiDARsim~\cite{manivasagam2020lidarsim} uses a multi-stage and data-driven approach to reconstruct the surfel-based~\cite{Pfister2000surfels} scene from real-world data and model the ray-drop property via a neural network for improving realism.
PCGen~\cite{li2023pcgen} utilizes the first peak averaging (FPA) ray-asting and surrogate model ray-drop to further close the domain gap.
Nevertheless, these explicit methods~\cite{manivasagam2020lidarsim,li2023pcgen,yang2020surfelgan} are sensitive to the geometry quality and only applicable to static scenes.
By leveraging generative models, some works~\cite{caccia2019deepgenerativemodelinglidar, zyrianov2022lidargen,zyrianov2024lidardm} provide an alternative but lack the physical control to model the real-world LiDAR sensors.

Several recent methods~\cite{tao2023lidar,Huang2023nfl,xue2024geonlf, yang2023unisim,zhang2024nerflidar} are proposed for high quality novel view LiDAR synthesis based on neural radiance fields (NeRF)~\cite{mildenhall2020nerf},
which combines the differentiable rendering capability of NeRF and the physical model of LiDAR sensors.
Notably, NFL~\cite{Huang2023nfl} demonstrates the detailed physical modeling of LiDAR sensors firstly, and provide a comprehensive reference for following works.
NeRF-LiDAR~\cite{zhang2024nerflidar} and UniSim~\cite{yang2023unisim} take multimodal inputs to enhance the performance of LiDAR re-simulation.
Furthermore, with the introduction of 4D hybrid features and additional tracking labels, some methods~\cite{Wu2023dynfl,zheng2024lidar4d,tonderski2023neurad,tang2024alignmif} extend to dynamic driving scenes.
However, all these NeRF-based approaches are constrained primarily by the high computational demands during training and rendering, and struggle to accurately model or re-simulate scenes with complex motions.

\PAR{Dynamic Scene Reconstruction.}
Expanding on the accomplishments of NeRF, some methods~\cite{park2021hypernerf,attal2023hyperreel,pumarola2020dnerf} build 4D neural representation to reconstruct object-level scenes via various extensions, such as continuous deformation fields~\cite{park2021nerfies} or spatial-temporal encoding~\cite{kplanes_2023}.
Some recent methods extend to dynamic urban scenes with extra supervision like optical flow~\cite{turki2023suds,ziyang2023snerf} and transformer features~\cite{yang2023emernerf}.
Apart from these works, another group of methods~\cite{ost2021nsg,wu2023mars,yang2023unisim,tonderski2023neurad} model the dynamic scene as the composition of dynamic objects and static background. UniSim~\cite{yang2023unisim} and NeuRAD~\cite{tonderski2023neurad} also simulate LiDAR observations at new views, which is similar to our work.
Nevertheless, these methods suffer from high computational costs on large scale scenes and cannot perform real-time rendering.
Aiming for rapid reconstruction and rendering, some methods prefer to represent the dynamic scene with 3D Gaussians~\cite{yan2024street, chen2023periodic,Zhou_2024_CVPR,zhou2023drivinggaussian}, which perform impressive results for novel view synthesis of camera sensors.
OmniRe~\cite{chen2024omnire} further constructs multiple local canonical spaces to model diverse dynamic objects in a driving log, including non-rigid pedestrians and cyclists.
Despite these advancements, there is a significant gap when they simulate other types of sensors like LiDAR, which is crucial for many perception and planning tasks.

\PAR{Gaussian Splatting.}
3D Gaussian Splatting~\cite{Kerbl20233dgs, luiten2023dynamic} made a breakthrough by using a differentiable splatting algorithm to render a scene represented by Gaussian primitives,
many works conduct extensions to 3DGS, such as anti-aliasing~\cite{Yu2023MipSplatting} and view-consistent rendering~\cite{Radl2024stopthepop}.
Based on the 3DGS, 2D Gaussian Splatting~\cite{Huang2DGS2024} collapses the 3D volume into a set of 2D oriented planar Gaussian disks and design a perspective-accurate 2D splatting process to model and reconstruct geometrically accurate radiance fields from multi-view images.
However, these rasterization-based rendering is difficult to model general ray effects(shadows, reflections etc.) and image formation processes that exhibit non-linear optical models, which limits its application for many tasks.
To address this issue, some approaches~\cite{blanc2024raygauss,R3DG2023,3dgrt2024,mai2024ever} propose to combine the ray tracing technologies with Gaussian particles.
3DGRT~\cite{3dgrt2024} utilizes the high-performance GPU ray tracing hardware to render the particle scenes in real-time.

\section{Method}
\label{sec:method}
In this section, we begin by delineating the problem formulation for the novel LiDAR view synthesis and the preliminary of Gaussian Splatting~\cite{Kerbl20233dgs}.
Following that, we provide a detailed exposition of our proposed \methodname.

\label{par:problem_formulation}
\PAR{Problem Formulation.}
In this problem, our input is a collection of LiDAR scans $\mathcal{L}$ captured by a moving sensor within a driving scenario,
along with the corresponding calibrated sensor poses $P$ and timestamps $T$.
Additionally, a group of bounding boxes $\mathcal{B}$ are provided to track the moving vehicles in the scene.

Our goal is to reconstruct this dynamic scene as an explicit representation and render realistic LiDAR views from any given moment $t_{new}$ and novel viewpoint $P_{new}$.
Furthermore, rendering must maintain high fidelity for various scene manipulations and sensor configurations.

\PAR{Gaussian Splatting.} Kerbl et al.~\cite{Kerbl20233dgs} propose to represent 3D scenes with 3D Gaussian primitives and render images using differentiable volume splatting. The $i$-th Gaussian is characterized by a full 3D covariance matrix $\boldsymbol{\Sigma}_{i}$ defined in world space~\cite{Zwicker2001ewa} centered at point (mean) $\mu_{i}$:
\begin{equation}
    \label{eq:gaussian}
    \begin{aligned}
        \mathcal{G}_{i}(\mathbf{x}) = \exp({-\frac{1}{2} (\mathbf{x}-\mu_{i})^\top \boldsymbol{\Sigma}_{i}^{-1}(\mathbf{x}-\mu_{i})}),
    \end{aligned}
\end{equation}
where the covariance matrix $\boldsymbol{\Sigma}_{i} = \mathbf{R}_{i} \mathbf{S}_{i} \mathbf{S}_{i}^\top \mathbf{R}_{i}^\top$ is factorized into a scaling matrix $\mathbf{S}_{i}$ and a rotation matrix $\mathbf{R}_{i}$.
Each Gaussian is paired with an opacity value $\sigma_{i}$ and the spherical harmonics coefficients (SHs)~\cite{Ravi2001sh} to model the view-dependent appearance $\mathbf{c}_{i}$.

When rendering novel views, 3D Gaussians are rasterized to image planes and form 2D Gaussians $\mathcal{G}^{'}$ with the 2D covariance $\boldsymbol{\Sigma}_{i}^{'} = \mathbf{J} \mathbf{W} \boldsymbol{\Sigma}_{i} \mathbf{W}^\top \mathbf{J}^\top$, where $\mathbf{W}$ is the viewing transformation and $\mathbf{J}$ is the Jacobian of the local affine transformation~\cite{Zwicker2001ewa}. The color of a pixel is calculated via alpha blending with depth ordered 2D Gaussians:
\begin{gather}
    \label{eq:volume_rendering}
    \mathcal{C}(\mathbf{x})  = \sum_{i=1}^{K} T_i \alpha_i \mathbf{c}_{i}, \\
    \begin{aligned}
        \text{where~~} \alpha_i  = \sigma_{i} \mathcal{G}_{i}^{'}(\mathbf{x}) \quad  \text{and} \quad  T_i=\prod_{j=1}^{i-1} (1 - \alpha_j).
    \end{aligned}
\end{gather}
\subsection{Overview of \methodname}
Towards the target of rendering realistic novel LiDAR views in real-time, we propose a novel framework named \methodname.
As illustrated in Fig.~\ref{fig:pipeline}, we represent the dynamic driving scene as a set of point clouds consisting of a static background and multiple foreground vehicles, each associated with Gaussian primitives (Section~\ref{sec:scene_representation}).
To ensure physical realism and high fidelity in rendering, we design a efficient and effective ray tracer based on Gaussian primitives to model the LiDAR imaging process (Section~\ref{sec:gaussian_ray_tracing}).
Moreover, we implement the backward pass of ray tracing to support differentiable LiDAR rendering (Section~\ref{sec:differentiable_lidar_rendering}) and scene optimization (Section~\ref{sec:optimization}).

\subsection{Dynamic Scene Representation}
\label{sec:scene_representation}

We decompose the dynamic scene into a static background and several moving vehicles, with each component represented by a distinct set of Gaussian primitives to softly learn the continuous scene geometry and radiance.
Similar to 3DGS~\cite{Kerbl20233dgs}, both the Gaussian primitives of background model and dynamic actor models share a set of common base parameters, including the mean position $\boldsymbol{\mu}$, covariance matrix $\boldsymbol{\Sigma}$, opacity $\sigma$, and SH coefficients,
the covariance matrix $\boldsymbol{\Sigma}$ can further be factorized into a scaling matrix $\mathbf{S}$ and a rotation matrix $\mathbf{R}$ as in Eq.~\ref{eq:gaussian}.

Additionally, we introduce learnable parameters on Gaussian primitives to emulate the intrinsic properties of LiDAR sensors, denoted by $\mathcal{P} = (\zeta, \beta)$, where $\zeta \in \mathbb{R}^{1}$ represents the reflection intensity and $\beta \in [0, 1]$ indicates the ray-drop probability.
Ray-drop is a common phenomenon of real-world LiDAR sensors that occurs when the return signal is too weak~\cite{man2023towards}, and the emitted ray is considered as dropped.
Unlike previous methods that directly model the ray-drop with a single variable~\cite{tao2023lidar,Huang2023nfl,Wu2023dynfl,zheng2024lidar4d}, we learn the ray-drop probability $\beta$ with two logits $(\beta_{\text{drop}}, \beta_{\text{hit}})$, activated by the softmax function \cite{berman2018lovasz}:
\begin{equation}
    \label{eq:ray_drop}
    \begin{aligned}
        \beta = \frac{e^{\beta_{\text{drop}}}}{e^{\beta_{\text{drop}}} + e^{\beta_{\text{hit}}}}.
    \end{aligned}
\end{equation}
Since the intensity and ray-drop are strongly affected by view directions, our method models $(\zeta, \beta_{\text{drop}}, \beta_{\text{hit}})$ with a set of SH coefficients as shown in Fig.~\ref{fig:pipeline}(a).

To handle the foreground moving vehicles, we utilize tracked bounding boxes to trace their motion trajectories.
Different from the background model, the mean position $\boldsymbol{\mu}_{o}$ and rotation matrix $\mathbf{R}_{o}$ of the Gaussian primitives for dynamic objects are defined within the object local coordinate system.
To transform them into the world coordinate system, same as~\cite{yan2024street,tonderski2023neurad}, we define the tracked poses for each rigid object as a series of rotation matrices $\{\mathbf{R}_{t}\}_{t=1}^{N_t} (\mathbf{R}_{t} \in \mathbb{R}^{3 \times 3})$ and translation vectors $\{\mathbf{T}_{t}\}_{t=1}^{N_t} (\mathbf{T}_{t} \in \mathbb{R}^{3 \times 1})$, where $N_t$ represents the number of frames.
The transformation is formulated as:
\begin{equation}
    \label{eq:object_pose_transform}
    \begin{aligned}
        \boldsymbol{\mu}_{w} & = \mathbf{R}_t \boldsymbol{\mu}_o + \mathbf{T}_t,
        \\
        \mathbf{R}_{w}       & = \mathbf{R}_t \mathbf{R}_o,
        \\
    \end{aligned}
\end{equation}
where $\boldsymbol{\mu}_{w}$ and $\mathbf{R}_{w}$ denote the mean position and rotation matrix of the Gaussian primitives in the world coordinate system, respectively.
With this representation, we can reconstruct and render the dynamic scene with easy composition of separate models, which further enables flexibly scene editing applications.

For the initialization of point clouds for background model, we estimate the normals of all Gaussian primitives with KNN algorithms and initialize the Gaussian orientations towards the normals.
Then we fuse the multi-frame LiDAR point clouds and downsample them via voxel downsampling with a voxel size of 0.15.
For object model less than $8K$ points, we randomly sample points inside the 3D bounding box and concatenate with the original points to reach $8K$ points.

\subsection{Gaussian-based Ray Tracing}
\label{sec:gaussian_ray_tracing}
To render novel LiDAR views, a straightforward approach is to rasterize the Gaussian primitives onto the LiDAR imaging plane and accumulated LiDAR properties through Eq.~\ref{eq:volume_rendering}, similar to the process of general camera sensors.
However, the vanilla rasterization approach of 3DGS~\cite{Kerbl20233dgs} does not support the rendering of cylindrical range image produced by LiDAR sensors.
Furthermore, the rasterization imaging process is not aligned with the active nature of LiDAR sensors, which rely on the emission and reception of laser beams.
Therefore, a ray-based rendering approach is essential for accurately simulating the LiDAR sensors.
Building upon these insights and inspired by the latest point-based ray tracing methodologies~\cite{3dgrt2024,blanc2024raygauss, R3DG2023, mai2024ever}, we develop a Gaussian-based ray tracer based on the NVIDIA OptiX framework~\cite{Parker2010optix} for hardware acceleration.

\begin{figure}[htbp]
    \centering
    \vspace{-2mm}
    \includegraphics[width=0.9\linewidth]{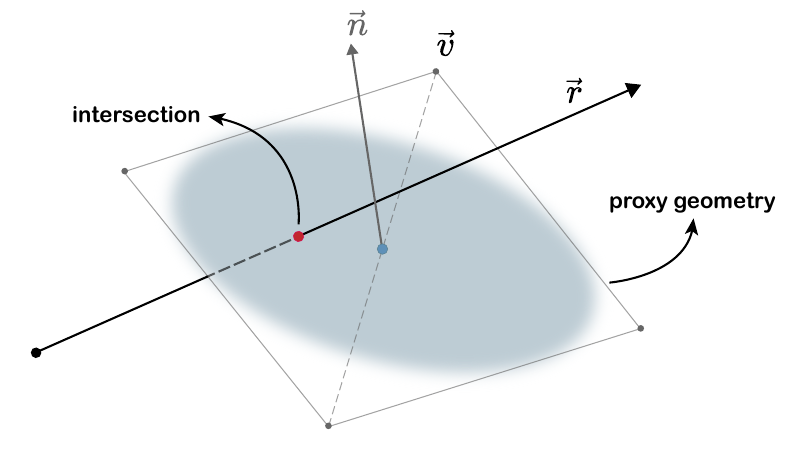}
    \caption{
        \textbf{Ray intersection with proxy geometries.} We construct the proxy geometry for each 2D Gaussian primitive as a pair of co-planar triangles, then the ray tracer performs intersection tests with the vertices $\vec{v}$ and surface normal $\vec{n}$.
    }
    \label{fig:ray_intersection}
    \vspace{-4mm}
\end{figure}

\PAR{Proxy Geometries.} In the realm of traditional graphics rendering~\cite{Lombardi21mvp,aliev2020neuralpointbasedgraphics}, proxy geometries serve as the simplified representations to approximate the complex original geometries, which can accelerate the ray tracing process.
Considering the efficiency in constructing the BVH and the performance of intersection tests, we opt for a two-dimensional Gaussian primitive in the form of a planar disk, as elaborated in~\cite{Huang2DGS2024}.
Thereby, the most efficient and simplest proxy geometry for encapsulating such Gaussian primitive is a pair of co-planar triangles (Fig.~\ref{fig:ray_intersection}).
Compared to axis-aligned bounding boxes (AABBs) or other polyhedron meshes,
this approach reduces the number of meshes and tightly wraps the Gaussian primitive.
Additionally, the sample locations are same as the ray intersections, without the approximation with maximum response, as described in~\cite{knoll2019efficient, 3dgrt2024}.
More details on proxy geometries for Gaussian variants can be found in our supplementary material.

\PAR{Forward Ray Tracing.}
After constructing the acceleration structures, our method proceed to generate batch rays from the LiDAR sensor, formulated as $\mathcal{R} = \{\mathbf{r}_{o}, \mathbf{r}_{d}\}$, where $\mathbf{r}_{o} \in \mathbb{R}^{N_{r} \times 3}$ and $\mathbf{r}_{d} \in \mathbb{R}^{N_{r} \times 3}$ denote the ray origins and normalized ray directions, with $N_{r}$ being the total number of batch rays.
our ray tracing pipeline is initialized by the \textit{optixLaunch}~\cite{Parker2010optix} program.
Specifically, given a ray $r$ within a batch of rays $\mathcal{R}$, the \textit{ray-gen} program is invoked to cast the ray against the BVH.
The tracer then utilizes triangle vertices and surface normal to compute intersections (Fig.~\ref{fig:ray_intersection}), using the \textit{any-hit} program to maintain the intersections in an sorted buffer.
However, sorting all intersections along each ray directly is inefficient, particularly in large-scale scenes with many Gaussian primitives.
To deal with this problem, we divide the ray into several chunks during the round of tracing, each containing a fixed number of intersections, thereby reducing the sorting overhead to a single chunk.
When the count of intersections reaches the chunk size, we take out the indices $\mathcal{I}$ of the intersected primitives and depth values $\mathcal{D}$ at the intersections from the sorted buffer.
Subsequently, we evaluate the Gaussian response and calculate the point-wise LiDAR properties $(\boldsymbol{\zeta}_{i}, \boldsymbol{\beta}_{i})$ at each sample point.
The pixel value is rendered through Eq.~\ref{eq:volume_rendering} by replacing $\mathbf{c}_{i}$ with $\zeta_{i}$ and $\beta_{i}$.
Following this, ray marching advances to the next chunk, starting from the last evaluated point.
This iterative process continues until all Gaussian primitives intersecting the ray have been traversed, or the accumulated transmittance along the ray falls below a predefined minimum threshold $T_{min}$.

\subsection{Differentiable LiDAR Rendering}
\label{sec:differentiable_lidar_rendering}

\PAR{LiDAR Modeling.} With the purpose of rendering LiDAR views, we present the LiDAR view as the form of range image.
Given a LiDAR scan $\mathcal{L}_{i}$ contains $\mathcal{M}_{i}$ points each parameterized by $(x, y, z, \zeta)$, we first calculate the distance $d$ to the sensor center with $d = \sqrt{x^2 + y^2 + z^2}$. Then the azimuth angle $\theta$ and elevation angle $\phi$ is computed as follows:
\begin{equation}
    \label{eq:azimuth_elevation}
    \begin{aligned}
        \theta & = \arctan(y, x),
        \\
        \phi   & = \arcsin(z, d).
    \end{aligned}
\end{equation}
Considering a LiDAR sensor with $\mathbf{H}$ laser beams in the vertical plane and $\mathbf{W}$ horizontal emissions, we can project the points onto the LiDAR imaging plane from the world coordinate system:
\begin{equation}
    \label{eq:range_image_projection}
    \begin{aligned}
        \binom{h}{w}=\binom{1-\left(\phi+|f_{down}| / f_v\right)}{(1-\theta / \pi) / 2} \cdot\binom{\mathbf{H}}{\mathbf{~W}},
    \end{aligned}
\end{equation}
where $(h, w)$ is the 2D coordinates on range image, and $f_v = | f_{up} | + | f_{down} |$ is the vertical FOV of LiDAR sensor.

\begin{figure*}[th]
    \centering
    \vspace{-4mm}
    \includegraphics[width=1.0\linewidth]{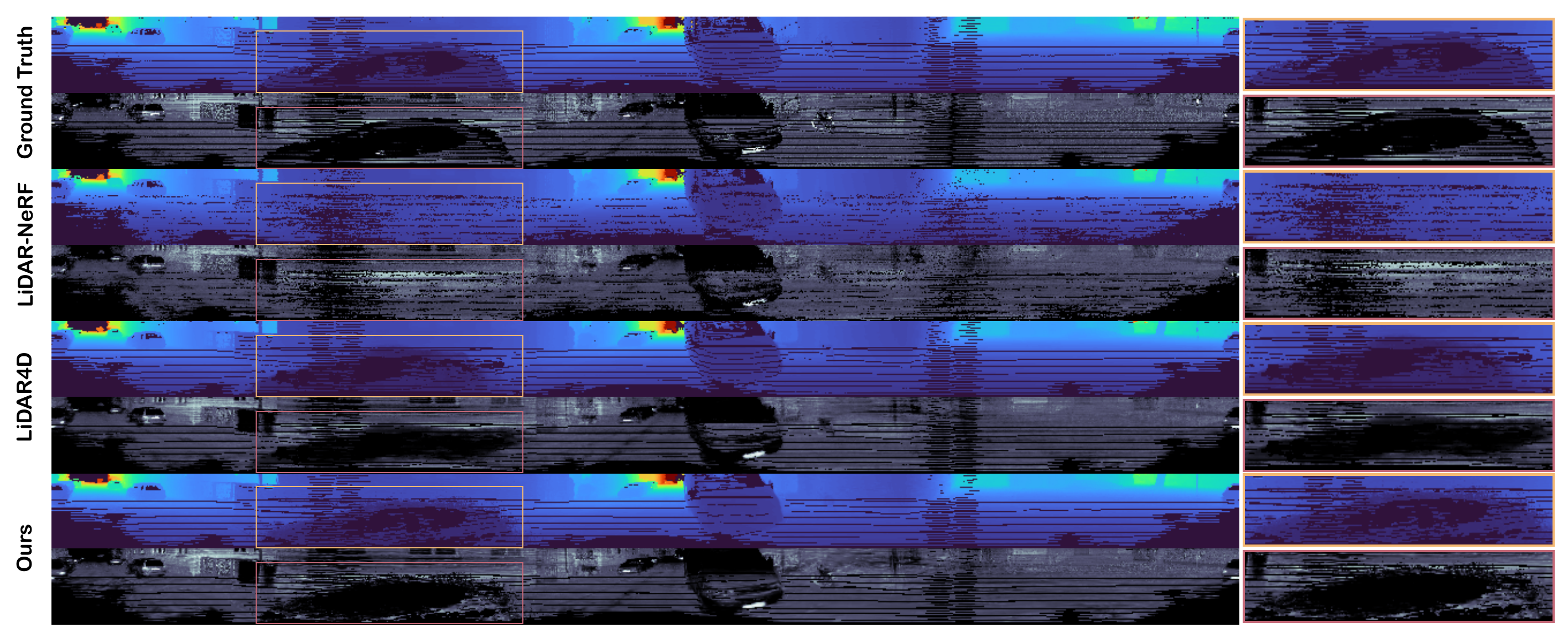}
    \vspace{-6mm}
    \caption{
        \textbf{Qualitative comparison on the KITTI-360~\cite{liao2022kitti} dataset.} Dynamic vehicles are zoomed in for better visualization.
    }
    \label{fig:nvs_kitti}
    \vspace{-4mm}
\end{figure*}

\PAR{Differentiable Rendering.}
Apart from the forward LiDAR rendering, our ray tracer also supports differentiable rendering to optimize the scene representation.
In previous works~\cite{Kerbl20233dgs, Huang2DGS2024}, the back-propagation is conducted in a back-to-front blending order:
\begin{equation}
    \label{eq:gaussian_backwards}
    \begin{aligned}
        \frac{\partial L}{\partial \alpha_i} = T_i \boldsymbol c_i - \sum^N_{j=i+1} T_j \alpha_j \boldsymbol c_j,
    \end{aligned}
\end{equation}
To compute gradients for the $i$-th Gaussian, we need the contributions from all subsequent blended Gaussians.
3DGS \cite{Kerbl20233dgs} meets this need by using per-tile sorted buffers.
However, this approach is not available for us due to the distinction between the panoramic LiDAR models and camera models.
A naive method to record the blending order in a global buffer during the forward pass is impractical, as it leads to unaffordable memory overhead and reduced ray tracing efficiency.
To address this issue, we take the strategy proposed by~\cite{Radl2024stopthepop} to perform the back-propagation in a front-to-back blending order, same as forward rendering:
\begin{equation}
    \label{raytracing_backwards}
    \begin{aligned}
        \frac{\partial L}{\partial \alpha_i} = T_i \boldsymbol c_i - (\boldsymbol{\mathcal C} - \mathcal C_i) / (1-\alpha_i),
    \end{aligned}
\end{equation}
where $\boldsymbol{\mathcal C}$ is the rendered pixel value and $\mathcal{C}_{i}$ is the accumulated attribute including the $i$-th Gaussian in front-to-back order.
Specifically, our backward pass starts by launching a new \textit{optixLaunch} program defined in the backwards kernel,
we then re-cast the same rays against the BVH to get the sorted intersections along the ray, computing the gradients via Eq.~\ref{raytracing_backwards} and accumulating them in shared global buffers with atomic operations.

\subsection{Optimization}
\label{sec:optimization}

\PAR{Loss Function.} Our method optimize the scene representation with the following loss functions:
\begin{equation}
    \label{eq:loss_function}
    \begin{aligned}
        \mathcal{L} = \mathcal{\lambda}_{d} \mathcal{L}_{d} + \mathcal{\lambda}_{i} \mathcal{L}_{i} + \mathcal{\lambda}_{r}\mathcal{L}_{r} + \mathcal{\lambda}_{\textit{CD}} \mathcal{L}_{\textit{CD}},
    \end{aligned}
\end{equation}
where the $\mathcal{\lambda}_{*}$ denote the weights of the individual loss terms
$\mathcal{L}_{*}$.
The depth loss $\mathcal{L}_{d}$ and intensity loss $\mathcal{L}_{i}$ follow $\mathcal{L}_{1}$ loss, and ray-drop loss $\mathcal{L}_{r}$ follows $\mathcal{L}_{bce}$ loss.
We employ the Chamfer Distance (\textit{CD}) loss $\mathcal{L}_{\textit{CD}}$~\cite{fan2016cd, zheng2024lidar4d} to jointly supervise scene geometry. The CD loss is calculated with the point-wise distance between the rendered and ground-truth point clouds:
\begin{equation}
    \label{cd_loss}
    \begin{aligned}
        \mathcal{L}_{\text{CD}} =\frac{1}{K} (\sum_{\hat{p}_i \in \hat{\mathcal{S}}} \min_{p_i \in S}\left\|\hat{p}_i-p_i\right\|_2^2 + \sum_{p_i \in S} \min_{\hat{p}_i \in \hat{\mathcal{S}}} \left\|p_i-\hat{p}_i\right\|_2^2),
    \end{aligned}
\end{equation}
where $K$ is $\min(|\hat{\mathcal{S}}|, |S|)$, $\hat{\mathcal{S}}$ and $S$ are the rendered and ground-truth point clouds, respectively.

\PAR{Ray-drop Refinement.}
In driving scenarios, we categorize the ray-drop effects into two types: \textit{scene-level} and \textit{sensor-level}.
\textit{Scene-level} ray-drop occurs due to environmental factors like reflective materials and long detection ranges.
While \textit{sensor-level} ray-drop is caused by the inherent sensor biases that are independent of the scene representation, such as the hardware noises and limited visibility.
Through our practice, we discovered that modeling ray-drops solely with SHs and optimizing them jointly with Gaussian primitives leads to inaccurate estimations in several areas.
Inspired by LiDAR4D~\cite{zheng2024lidar4d}, we utilize a UNet~\cite{Navab2015unet} $\mathcal{F}_{\text{refine}}$ to refine the \textit{sensor-level} ray-drop effects after the Gaussian optimization.
Contrary to LiDAR4D, we feed the ray origins and ray directions as additional inputs so that the UNet can better percept the spatial movement of the sensor:
\begin{equation}
    \label{eq:raydrop_refine}
    \begin{aligned}
        \beta_{\text{refine}} = \mathcal{F}_{\text{refine}}(d, \zeta, \beta, \mathbf{r}_{o}, \mathbf{r}_{d}).
    \end{aligned}
\end{equation}
We refine the rendered ray-drop mask via binary cross-entropy loss as follows:
\begin{equation}
    \label{eq:raydrop_loss}
    \begin{aligned}
        \mathcal{L}_r =
        \frac{1}{|\mathcal{R}|} \displaystyle\sum_{\mathbf{r} \in \mathcal{R}} \mathcal{L}_{bce}(\hat{P}(\mathbf{r}), P(\mathbf{r})).
    \end{aligned}
\end{equation}
Moreover, we apply a random horizontal rotation to the sampled sensor pose and ground-truth image to avoid the overfitting to the specific pose.
\begin{table*}[th]
    \centering
    \vspace{-2mm}
    \caption{\textbf{Quantitative comparison on the Waymo Open Dataset~\cite{sun2020waymo}}. The resolution of range image is 64×2650 and metrics are averaged over all sequences.
        The cell colors present the \colorbox{colorfirst}{best} and the \colorbox{colorsecond}{second best} results, respectively.}
    \vspace{-0.2cm}
    \resizebox{\textwidth}{!}{
        \renewcommand{\arraystretch}{1.25}
        \begin{tabular}{c|cc|ccccc|ccccc|cc}
            \toprule \multirow[c]{2}{*}{ Method } & \multicolumn{2}{c|}{ Efficiency} & \multicolumn{5}{c|}{ Depth } & \multicolumn{5}{c|}{ Intensity } & \multicolumn{2}{c}{ Point Cloud }                                                                                                                                                                                                                     \\
                                                  & FPS                              & Storage                      & RMSE$\downarrow$                 & MedAE$\downarrow$                 & LPIPS$\downarrow$  & SSIM$\uparrow$     & PSNR$\uparrow$      & RMSE$\downarrow$   & MedAE$\downarrow$  & LPIPS$\downarrow$  & SSIM$\uparrow$     & PSNR$\uparrow$      & CD$\downarrow$     & F-score$\uparrow$  \\
            \midrule
            LiDAR-NeRF~\cite{tao2023lidar}        & \cellsecond 0.98                 & \cellsecond 1.6 GB           & 7.7258                           & 0.0518                            & 0.3414             & 0.6817             & 20.5204             & 0.0659             & 0.0108             & 0.1893             & 0.7492             & 23.1612             & 0.1815             & 0.9184             \\
            DyNFL~\cite{Wu2023dynfl}              & 0.21                             & 14.9 GB                      & 6.9787                           & 0.0388                            & \cellsecond 0.3010 & \cellsecond 0.7080 & 21.3094             & 0.0662             & \cellsecond 0.0093 & \cellfirst 0.1382  & 0.7555             & \cellsecond 24.0473 & 0.1182             & 0.7786             \\
            LiDAR4D~\cite{zheng2024lidar4d}       & 0.17                             & 7.7 GB                       & \cellsecond 6.6234               & \cellsecond 0.0379                & 0.3406             & 0.7008             & \cellsecond 21.8413 & \cellsecond 0.0634 & \cellfirst 0.0086  & 0.1854             & \cellsecond 0.7756 & \cellfirst 24.0838  & \cellsecond 0.1060 & \cellsecond 0.9437 \\
            \textbf{Ours}                         & \cellfirst 20.1                  & \cellfirst 1.37 GB           & \cellfirst 6.4577                & \cellfirst 0.0340                 & \cellfirst 0.2918  & \cellfirst 0.7330  & \cellfirst 22.1581  & \cellfirst 0.0627  & 0.0101             & \cellsecond 0.1652 & \cellfirst 0.7828  & 23.9079             & \cellfirst 0.1002  & \cellfirst 0.9458  \\

            \bottomrule
        \end{tabular}}
    \label{tab:nvs_waymo}
\end{table*}

\begin{table*}[th]
    \centering
    \caption{\textbf{Quantitative comparison on the KITTI-360 benchmarks~\cite{liao2022kitti}}.
        The resolution of range image is 66×1030 and metrics are averaged over all sequences.
        The color annotations are consistent with the Table~\ref{tab:nvs_waymo} above.}
    \vspace{-0.2cm}
    \resizebox{\textwidth}{!}{
        \renewcommand{\arraystretch}{1.25}
        \begin{tabular}{c|cc|ccccc|ccccc|cc}
            \toprule \multirow[c]{2}{*}{ Method }   & \multicolumn{2}{c|}{ Efficiency} & \multicolumn{5}{c|}{ Depth } & \multicolumn{5}{c|}{ Intensity } & \multicolumn{2}{c}{ Point Cloud }                                                                                                                                                                                                                     \\
                                                    & FPS                              & Storage                      & RMSE$\downarrow$                 & MedAE$\downarrow$                 & LPIPS$\downarrow$  & SSIM$\uparrow$     & PSNR$\uparrow$      & RMSE$\downarrow$   & MedAE$\downarrow$  & LPIPS$\downarrow$  & SSIM$\uparrow$     & PSNR$\uparrow$      & CD$\downarrow$     & F-score$\uparrow$  \\
            \midrule
            PCGen~\cite{li2023pcgen}                & 0.05                             & 5.23 GB                      & 5.6853                           & 0.2040                            & 0.5391             & 0.4903             & 23.1675             & 0.1970             & 0.0763             & 0.5926             & 0.1351             & 14.1181             & 0.4636             & 0.8023             \\
            LiDARsim~\cite{manivasagam2020lidarsim} & 0.9                              & \cellsecond 0.76 GB          & 6.9153                           & 0.1279                            & 0.2926             & 0.6342             & 21.4608             & 0.1666             & 0.0569             & 0.3276             & 0.3502             & 15.5853             & 3.2228             & 0.7157             \\
            LiDAR-NeRF~\cite{tao2023lidar}          & \cellsecond 1.8                  & 1.61 GB                      & 4.0886                           & 0.0556                            & 0.2712             & 0.6309             & 26.0590             & 0.1464             & 0.0438             & 0.3590             & 0.3567             & 16.7621             & 0.1502             & 0.9073             \\
            LiDAR4D~\cite{zheng2024lidar4d}         & 0.4                              & 7.38 GB                      & \cellsecond 3.5256               & \cellfirst 0.0404                 & \cellsecond 0.1051 & \cellsecond 0.7647 & \cellsecond 27.4767 & \cellsecond 0.1195 & \cellsecond 0.0327 & \cellsecond 0.1845 & \cellsecond 0.5304 & \cellsecond 18.5561 & \cellsecond 0.1089 & \cellfirst 0.9272  \\
            \textbf{Ours}                           & \cellfirst 42.7                  & \cellfirst 0.41 GB           & \cellfirst 3.4671                & \cellsecond 0.0512                & \cellfirst 0.1016  & \cellfirst 0.8406  & \cellfirst 27.6755  & \cellfirst 0.1115  & \cellfirst 0.0271  & \cellfirst 0.1812  & \cellfirst 0.6077  & \cellfirst 19.0862  & \cellfirst 0.1077  & \cellsecond 0.9255 \\

            \bottomrule
        \end{tabular}}
    \label{tab:nvs_kitti360}
\end{table*}

\section{Implementation details}
\label{sec:implementation_details}
The implementation of \methodnameblank is based on PyTorch~\cite{paszke2019pytorch} and our custom CUDA kernels~\cite{Ghorpade2012cuda} with NVIDIA OptiX framework~\cite{Parker2010optix}.
We train \methodnameblank for 30000 iterations with Adam optimizers~\cite{kingma2014adam} following the configurations of 3DGS~\cite{Kerbl20233dgs}.
After that, we refine the UNet~\cite{Navab2015unet} model for 500 epochs with a learning rate of $1e^{-3}$.
In practice, the loss weights $\lambda_{d}$, $\lambda_{i}$, $\lambda_{r}$, and $\lambda_{\textit{CD}}$ are set to 0.1, 0.1, 0.01, 0.01 respectively.
During ray tracing, we set the chunk size to 16 for the trade-off between memory consumption and efficiency. The near plane is set to 0.2.
All the experiments are conducted on one single RTX 4090 GPU.

\section{Experiments}
\label{sec:experiments}

\subsection{Experimental Setup}
\label{sec:experimental_setup}

\PAR{Datasets.}
We conducted experiments on the public Waymo Open Dataset~\cite{sun2020waymo} and KITTI-360 benchmarks~\cite{liao2022kitti}.
Both datasets is equipped with a 64-beam LiDAR sensor and an acquisition frequency of 10Hz.
On Waymo dataset, we follow the settings of DyNFL~\cite{Wu2023dynfl}, selecting 4 dynamic sequences and 4 static sequences for experiments.
Each selected sequence has 50 continuous frames, we sample every $10^{th}$ frame in the sequence as the test frames and use the remaining for training, the resolution of the range image is 64×2650.
On KITTI-360 dataset, we select 6 dynamic sequences and 4 static sequences for experiments and evaluate our method with the settings of LiDAR4D~\cite{zheng2024lidar4d}. The sample strategy is as same as the Waymo dataset.
However, the KITTI-360 dataset only provides per-frame point clouds instead of the raw range images, so we fuse the multi-frame point clouds and project the fused point clouds to range images with the resolution of 66×1030.

\PAR{Baseline Methods.}
We perform a comprehensive comparison of our method with various types of baselines:
(1) LiDARsim~\cite{manivasagam2020lidarsim} and PCGen~\cite{li2023pcgen} are mesh-based reconstruction methods, we reproduce the results under the same settings in their papers.
(2) LiDAR-NeRF~\cite{tao2023lidar} is the first NeRF-based method for LiDAR re-simulation, we use the official implementation for comparison.
(3) LiDAR4D~\cite{zheng2024lidar4d} and DyNFL~\cite{Wu2023dynfl} extend neural fields to dynamic scenes through 4D hybrid features and tracked bounding boxes, which are the primary methods we compare with.
Please refer to the supplementary materials for more implementation details of baseline methods.

\PAR{Evaluation Metrics.}
Following the previous works~\cite{tao2023lidar, Huang2023nfl,zheng2024lidar4d,Wu2023dynfl}, we evaluate the performance of our method with following metrics:
We use Chamfer Distance (CD)~\cite{fan2016cd} and F-score (error threshold is 5cm) to quantify the 3D geometric error between the generated and ground-truth point clouds.
For evaluating range accuracy, we report the Root Mean Square Error (RMSE) and Median Absolute Error (MedAE), along with the PSNR, SSIM~\cite{Wang2004ssim}, and LPIPS~\cite{Zhang_2018_CVPR} to measure the overall variance.
The intensity results are evaluated in the same manner with the range.
In addition, we present the efficiency of our method in terms of the inference FPS and training GPU storage.

\begin{figure*}[th]
    \centering
    \vspace{-2mm}
    \includegraphics[width=1.0\linewidth]{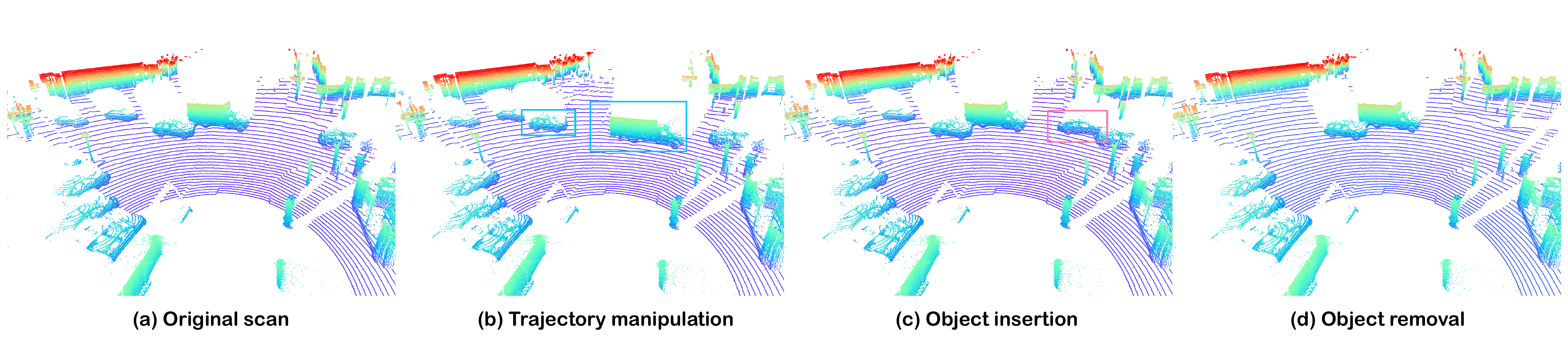}
    \caption{
        \textbf{Scene editing results on the Waymo dataset~\cite{sun2020waymo}.} Our method supports various scene editing operations, including trajectory manipulation, object insertion, and object removal. The boxes ($\textcolor{cyan}{\Box} / \textcolor{pink}{\Box}$) highlight the edited objects.
    }
    \label{fig:scene_editing}
    \vspace{-2mm}
\end{figure*}

\begin{table}[t]
    \centering
    \caption{\textbf{Ablation studies on dynamic sequence of the KITTI-360 dataset~\cite{liao2022kitti}.} $\mathcal{L}_{\text{CD}}$: Chamfer Distance loss, $\mathcal{I}$: Point cloud initialization with normals, $\mathcal{M}$: Ray-drop modeling.
    Sec.~\ref{sec:ablations} provides a detailed description for each ablation term.}
    \vspace{-0.2cm}
    \resizebox{0.48\textwidth}{!}{
        \begin{tabular}{c|ccc|ccc|cc}
            \toprule \multirow[c|]{2}{*}{}         & \multicolumn{3}{c|}{ Depth } & \multicolumn{3}{c|}{ Intensity } & \multicolumn{2}{c}{ Point Cloud }                                                                                                      \\
                                                   & RMSE$\downarrow$             & SSIM$\uparrow$                   & PSNR$\uparrow$                    & RMSE$\downarrow$  & SSIM$\uparrow$    & PSNR$\uparrow$     & CD$\downarrow$    & F-score$\uparrow$ \\
            \midrule
            \textit{w/o} $\mathcal{L}_{\text{CD}}$ & 3.5053                       & 0.8393                           & 27.5526                           & 0.1120            & 0.6068            & 19.0553            & 0.1192            & 0.9464            \\
            \textit{w/o} $\mathcal{I}$             & 4.0597                       & 0.8343                           & 26.2554                           & 0.1141            & 0.6024            & 18.8896            & 0.1363            & 0.9389            \\
            \textit{w/o} $\mathcal{M}$             & 3.7445                       & 0.7989                           & 26.9473                           & 0.1179            & 0.5687            & 18.6051            & 0.1234            & 0.9286            \\
            \textbf{Ours}                          & \cellfirst 3.4671            & \cellfirst 0.8406                & \cellfirst 27.6755                & \cellfirst 0.1115 & \cellfirst 0.6077 & \cellfirst 19.0862 & \cellfirst 0.1077 & \cellfirst 0.9255 \\

            \bottomrule
        \end{tabular}}
    \vspace{-6mm}
    \label{tab:ablation_all}
\end{table}

\subsection{Comparisons with the State-of-the-art}
\label{sec:comparisons}

Tabs.~\ref{tab:nvs_waymo},~\ref{tab:nvs_kitti360} and Fig.~\ref{fig:nvs_kitti} show the quantitative and qualitative comparisons with the SOTA NeRF-based methods~\cite{tao2023lidar,Wu2023dynfl,zheng2024lidar4d} on the Waymo~\cite{sun2020waymo} and KITTI-360~\cite{liao2022kitti} datasets.
As evident in Tab.~\ref{tab:nvs_waymo} and Tab.~\ref{tab:nvs_kitti360}, our method achieves notably better rendering quality compared to the SOTA methods~\cite{Wu2023dynfl,zheng2024lidar4d},
while rendering at two orders of magnitude faster and requiring significantly less training storage.
As shown in Fig.~\ref{fig:nvs_kitti}, LiDAR-NeRF~\cite{tao2023lidar} fails to model the dynamic objects,
while LiDAR4D~\cite{zheng2024lidar4d} suffers from blurry results and ray-drop estimation errors due to the lack of capacity of their model to capture the complex scenes.
In contrast, our \methodnameblank is capable of generating high-fidelity and physically plausible LiDAR views for driving scenarios.
More comparison results on the Waymo dataset~\cite{sun2020waymo} can be found in supplementary materials.

\subsection{Ablations Studies}
\label{sec:ablations}

We perform ablation studies on the dynamic sequence of the KITTI-360 dataset~\cite{liao2022kitti}.
More qualitative and quantitative results of the ablation studies are provided in our supplementary materials.

\PAR{Ablation study on loss functions.}
As shown in Tab.~\ref{tab:ablation_all} row 1, remove the $\mathcal{L}_{\text{CD}}$ terms not only reduces the scene geometric accuracy (CD and F-score), but also leads to the degradation of other performance metrics.

\PAR{Ablation study on point cloud initialization with normals.}
The "\textit{w/o} $\mathcal{I}$" variant removes the normal-guided initialization for LiDAR point clouds (Sec.~\ref{sec:scene_representation}), which significantly impedes the quality of the scene geometry.

\PAR{Ablation study on ray-drop modeling.}
The "\textit{w/o} $\mathcal{M}$" variant removes the proposed ray-drop modeling module (Eq.~\ref{eq:ray_drop}) and replaces it with a simple ray-drop estimation.
As shown in Tab.~\ref{tab:ablation_all}, this variant leads to the ray-drop estimation errors and inferior rendering quality.

\subsection{Applications}
\label{sec:applications}

\PAR{Scene editing.}
Our instance-aware scene representation (Sec.~\ref{sec:scene_representation}) models each component separately and enables a wide range of scene editing operations.
Fig.~\ref{fig:scene_editing} demonstrates the high-fidelity scene editing results on the Waymo dataset~\cite{sun2020waymo}, including trajectory manipulation (b), object insertion (c), and object removal (d).

\begin{figure}[th]
    \centering
    \includegraphics[width=1.0\linewidth]{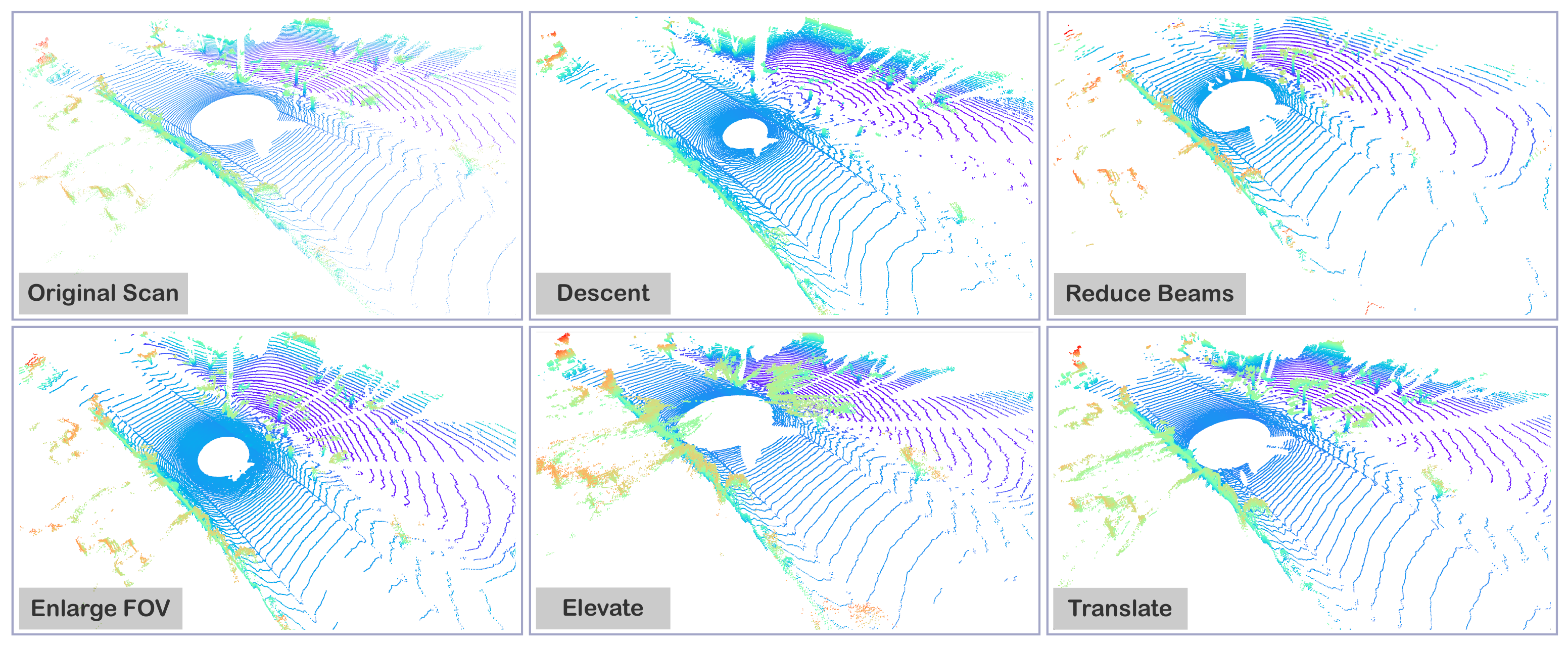}
    \caption{
        \textbf{LiDAR re-simulation results with different sensor configurations on the Waymo dataset~\cite{sun2020waymo}.}
        Our \methodnameblank can generate realistic LiDAR point clouds with varying sensor poses, beam numbers, and FOV settings.
    }
    \label{fig:sensor_configuration}
    \vspace{-2mm}
\end{figure}

\PAR{Sensor re-simulation.}
Thanks to our effective Gaussian-based ray tracer (Sec.~\ref{sec:gaussian_ray_tracing}), we can easily re-simulate LiDAR sensors with varying configurations.
As shown in Fig.~\ref{fig:sensor_configuration}, our method can generate realistic LiDAR point clouds with different sensor poses, beam numbers and FOV settings.
This flexibility is particularly beneficial for downstream tasks in autonomous driving.

\section{Conclusion and Discussion}
\label{sec:conclusion}
In this paper, we propose \methodname, a novel framework designed for real-time, high-fidelity LiDAR re-simulation of dynamic driving scenarios.
We decompose dynamic scenes into a static background and multiple moving objects, each represented by a set of Gaussian primitives, and incorporate scene graphs to handle scene dynamics.
Furthermore, we develope a differentiable LiDAR rendering pipeline that utilizes the hardware-accelerated ray tracing algorithms to generate realistic novel LiDAR views, also maintaining great performance under various scene editing operations and sensor configurations.
Extensive experiments across a variety of public autonomous driving benchmarks have demonstrated that our method achieves state-of-the-art performance in terms of both rendering quality and efficiency.

However, our method has certain limitations.
Firstly, \methodnameblank cannot accurately model non-rigid objects such as pedestrians and cyclists due to their substantial deformations across frames, which are important for some downstream tasks.
Moreover, the performance and rendering speed of \methodnameblank are impacted when dealing with long driving sequences, as the number of Gaussian primitives increases dramatically.
How to model non-rigid objects and improve the efficiency for long driving sequences are remaining challenges for future works.

{
    \small
    \bibliographystyle{ieeenat_fullname}
    \bibliography{main}
}

\newpage

\twocolumn[
    \centering
    \Large
    \textbf{\thetitle}\\
    \vspace{0.5em}Supplementary Material \\
    \begin{center}
    \captionsetup{type=figure}
    \vspace{2mm}
    \includegraphics[width=1.0\linewidth]{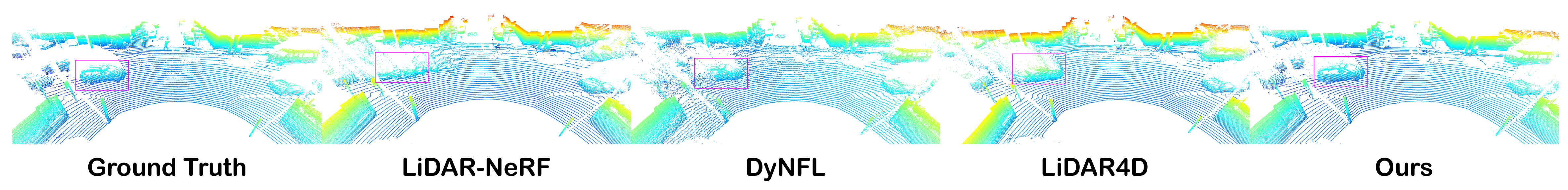}
    \captionof{figure}{
        \textbf{Qualitative comparison of novel view LiDAR point clouds on Waymo Open Dataset~\cite{sun2020waymo}.}
        Our \methodnameblank generates a realistic novel LiDAR view with accurate scene geometry and high-frequency details of dynamic objects.
    }
    \label{fig:supp_pcd_waymo}
\end{center}

    \bigbreak
] %

\appendix

\tolerance=40
In this supplementary material, we begin by presenting additional implementation details about the evaluation datasets and the baseline methods in Sec.~\ref{sec:implementation}.
Then we display the further ablation studies in Sec.~\ref{sec:suppl_ablations} and comparison results on Waymo~\cite{sun2020waymo} and KITTI-360~\cite{liao2022kitti} datasets in Sec.~\ref{sec:exp_results}.
Subsequently, Sec.~\ref{sec:suppl_applications} showcases more applications of our method.

\section{Implementation details}
\label{sec:implementation}

\subsection{Evaluation datasets.}
\label{sec:datasets}
We evaluate our method on the Waymo Open Dataset~\cite{sun2020waymo} and the KITTI-360 dataset~\cite{liao2022kitti}.
Following previous works~\cite{tao2023lidar,zheng2024lidar4d,Wu2023dynfl}, we select the static and dynamic sequences from the both datasets.
The specific selected sequence names and corresponding ids are listed in Tab.~\ref{tab:dataset_seqs_kitti360} and Tab.~\ref{tab:dataset_seqs_waymo}.

\begin{table}[ht]
    \centering
    \vspace{-0.2cm}
    \caption{\textbf{The selected sequences from the KITTI-360~\cite{liao2022kitti} dataset for evaluation.} $\mathcal{S}$ and $\mathcal{D}$ denote static and dynamic sequences, respectively.}
    \vspace{-0.1cm}
    \resizebox{0.48\textwidth}{!}{
        \renewcommand{\arraystretch}{0.8}
        \begin{tabular}{ccccc}
            \toprule
            Scene name   & Type          & Sequence id     & Start frame & End frame \\
            \midrule
            \textit{ks1} & $\mathcal{S}$ & Seq 1538-1601   & 1538        & 1601      \\
            \textit{ks2} & $\mathcal{S}$ & Seq 1728-1791   & 1728        & 1791      \\
            \textit{ks3} & $\mathcal{S}$ & Seq 1908-1971   & 1908        & 1971      \\
            \textit{ks4} & $\mathcal{S}$ & Seq 3353-3415   & 3353        & 3415      \\
            \textit{kd1} & $\mathcal{D}$ & Seq 2351-2400   & 2351        & 2400      \\
            \textit{kd2} & $\mathcal{D}$ & Seq 4951-5000   & 4951        & 5000      \\
            \textit{kd3} & $\mathcal{D}$ & Seq 8121-8170   & 8121        & 8170      \\
            \textit{kd4} & $\mathcal{D}$ & Seq 10201-10250 & 10201       & 10250     \\
            \textit{kd5} & $\mathcal{D}$ & Seq 10751-10800 & 10751       & 10800     \\
            \textit{kd6} & $\mathcal{D}$ & Seq 11401-11450 & 11401       & 11450     \\
            \bottomrule
        \end{tabular}}
    \vspace{-4mm}
    \label{tab:dataset_seqs_kitti360}
\end{table}

\begin{table}[ht]
    \centering
    \caption{\textbf{The selected sequences from the Waymo Open~\cite{sun2020waymo} dataset for evaluation.} $\mathcal{S}$ and $\mathcal{D}$ denote static and dynamic sequences, respectively.}
    \vspace{-0.1cm}
    \resizebox{0.48\textwidth}{!}{
        \renewcommand{\arraystretch}{0.8}
        \begin{tabular}{cccccc}
            \toprule
            Scene name   & Type          & Sequence id & Start frame & End frame \\
            \midrule
            \textit{ws1} & $\mathcal{S}$ & Seg 113792  & 1           & 50        \\
            \textit{ws2} & $\mathcal{S}$ & Seg 106762  & 1           & 50        \\
            \textit{ws3} & $\mathcal{S}$ & Seg 177619  & 1           & 50        \\
            \textit{ws4} & $\mathcal{S}$ & Seg 117240  & 1           & 50        \\
            \textit{wd1} & $\mathcal{D}$ & Seg 108305  & 148         & 197       \\
            \textit{wd2} & $\mathcal{D}$ & Seg 132712  & 51          & 100       \\
            \textit{wd3} & $\mathcal{D}$ & Seg 100721  & 1           & 50        \\
            \textit{wd4} & $\mathcal{D}$ & Seg 105003  & 148         & 197       \\
            \bottomrule
        \end{tabular}}
    \vspace{-4mm}
    \label{tab:dataset_seqs_waymo}
\end{table}

\subsection{Baseline methods}
\label{baselines}

\PAR{LiDARsim and PCGen.}
LiDARsim~\cite{manivasagam2020lidarsim} and PCGen~\cite{li2023pcgen} are surfel-based reconstruction methods.
Since the official implementation is not publicly available, we re-implement these two methods based on the codebase provided by the LiDAR-NeRF~\cite{tao2023lidar} and follow the same experimental settings on the KITTI-360 dataset~\cite{liao2022kitti}.

\PAR{LiDAR-NeRF.}
LiDAR-NeRF~\cite{tao2023lidar} is the first NeRF-based method for LiDAR re-simulation, we directly adopt the official implementation.
For KITTI-360 dynamic sequences and Waymo~\cite{sun2020waymo} scenes, we adjust the scene scales and LiDAR resolutions for fair comparison.

\PAR{LiDAR4D.}
LiDAR4D~\cite{zheng2024lidar4d} utilizes a 4D hybrid representation combined with multi-planar and grid features for LiDAR re-simulation.
We adopt the official implementation and follow the same experimental settings as their paper on the KITTI-360 dataset.
For the Waymo dataset, we preprocess the dataset following the same procedure as LiDAR4D and adjust the LiDAR resolutions.
The ray-drop refinement is also conducted for evaluation sequences.

\PAR{DyNFL.}
DyNFL~\cite{Wu2023dynfl} leverages the bounding boxes of moving objects to construct an editable neural field for high-fidelity re-simulation of LiDAR scans.
We follow the original implementation based on NFL Studio~\cite{Huang2023nfl} and the settings for the Waymo dataset.

\begin{figure*}[th]
    \centering
    \includegraphics[width=1.0\linewidth]{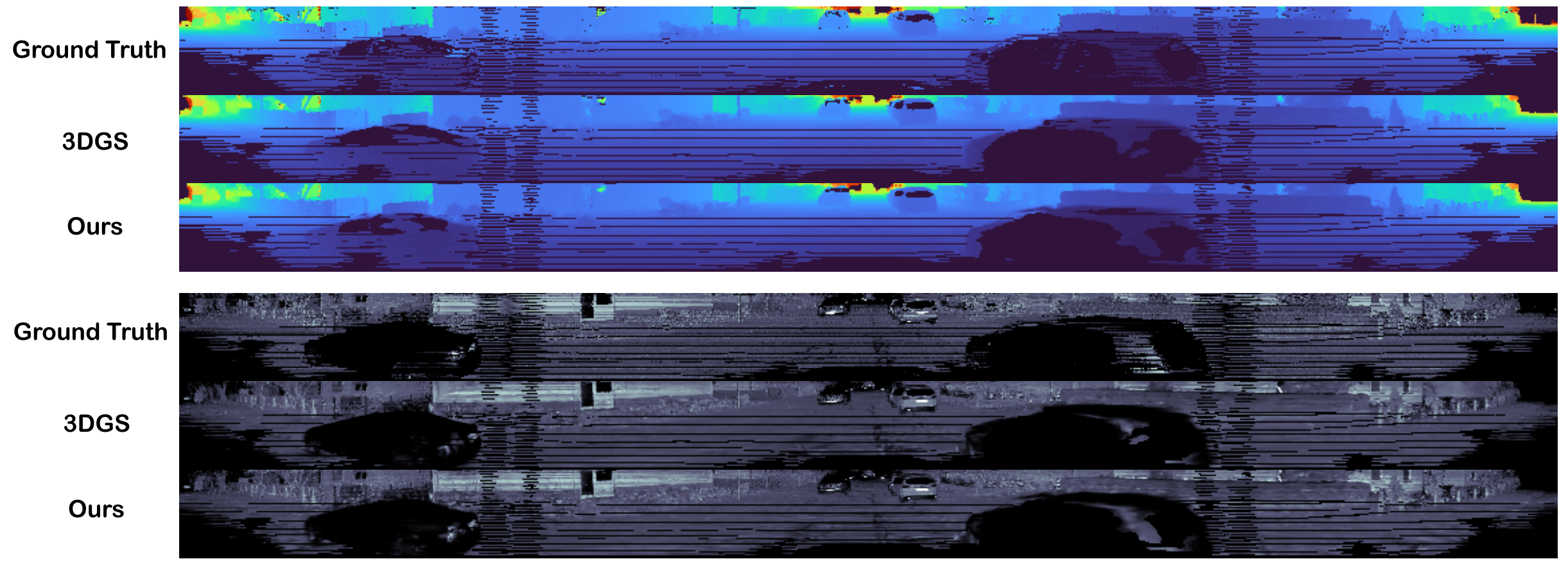}
    \caption{
        \textbf{Qualitative results of ablation study on ray tracing with Gaussian variants}.
    }
    \label{fig:supp_ablation_gs}
\end{figure*}

\begin{figure*}[th]
    \centering
    \includegraphics[width=1.0\linewidth]{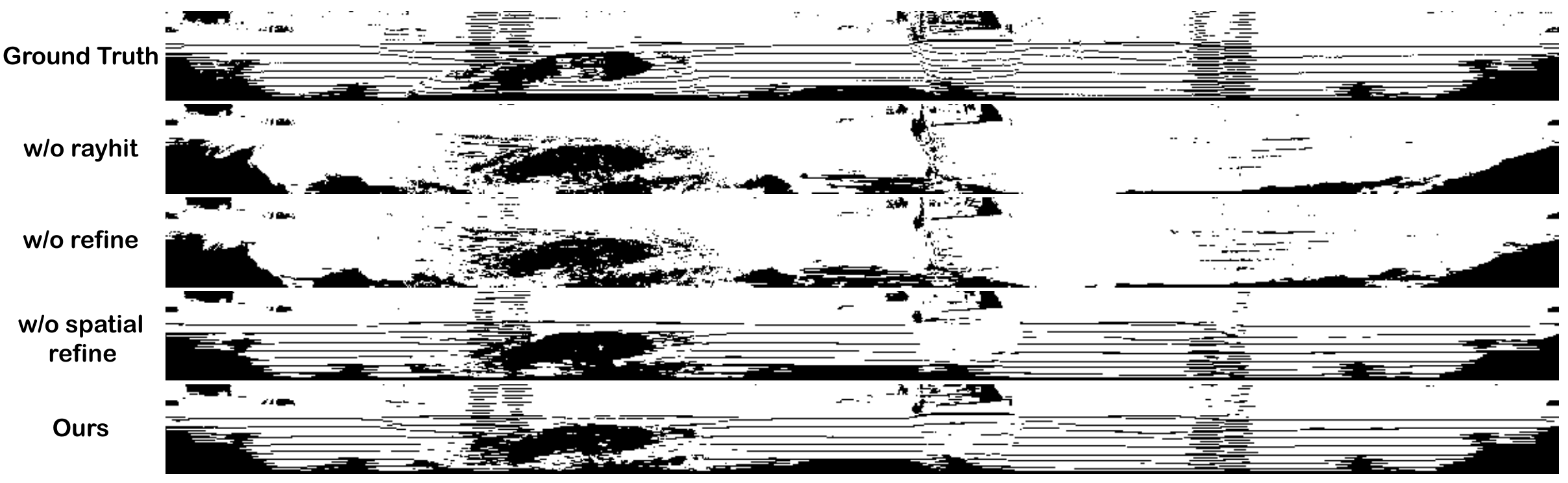}
    \caption{
        \textbf{Qualitative results of ablation study on ray-drop modeling and refinement}.
    }
    \label{fig:supp_ablation_raydrop}
\end{figure*}

\subsection{Gaussian Densification.}
We adopt the adaptive control techniques from 3DGS~\cite{Kerbl20233dgs} during optimization, which includes operations such as pruning, cloning, and splitting.
However, unlike the vanilla 3DGS~\cite{Kerbl20233dgs}, which tracks screen-space gradients of particle positions for cloning and splitting decisions, our approach utilizes gradients in 3D world-space.
This method is more general and suitable in our ray tracing context since the forward and backward passes are performed in 3D space.
Furthermore, to prevent object Gaussians from expanding into occluded areas, we follow the strategy of~\cite{yan2024street} and sample a set of points for each object model to form a probability distribution function.
During optimization, Gaussians associated with sampled points that fall outside the bounding box are pruned to avoid excessive growth.

\begin{table}[ht]
    \centering
    \caption{\textbf{Quantitative results of ablation study on ray tracing with Gaussian variants}.
        The cell colors present the \colorbox{colorfirst}{best} and the \colorbox{colorsecond}{second best} results, respectively.}
    \vspace{-0.1cm}
    \resizebox{0.48\textwidth}{!}{
        \renewcommand{\arraystretch}{1.0}
        \begin{tabular}{ccccccc}
            \toprule
            Method                       & FPS$\uparrow$  & RMSE$\downarrow$   & LPIPS$\downarrow$  & PSNR$\uparrow$      & CD$\downarrow$     & F-score$\uparrow$  \\
            \midrule
            3D Gaussians                 & \cellsecond 29 & \cellsecond 3.6716 & \cellsecond 0.1145 & \cellsecond 27.0976 & \cellsecond 0.3553 & \cellsecond 0.8899 \\
            \textbf{2D Gaussians (Ours)} & \cellfirst 42  & \cellfirst 3.4671  & \cellfirst 0.1070  & \cellfirst 27.6755  & \cellfirst 0.1077  & \cellfirst 0.9255  \\

            \bottomrule
        \end{tabular}}
    \vspace{-2mm}
    \label{tab:ablation_gs}
\end{table}

\begin{table}[ht]
    \centering
    \caption{\textbf{Quantitative results of ablation study on ray-drop modeling and refinement}.
        The cell colors present the \colorbox{colorfirst}{best} and the \colorbox{colorsecond}{second best} results, respectively.}
    \vspace{-0.1cm}
    \resizebox{0.48\textwidth}{!}{
        \renewcommand{\arraystretch}{1.0}
        \begin{tabular}{cccccc}
            \toprule
            Method                                      & RMSE$\downarrow$   & LPIPS$\downarrow$  & PSNR$\uparrow$      & CD$\downarrow$     & F-score$\uparrow$  \\
            \midrule
            \textit{w/o} $\mathcal{R}_{\text{hit}}$     & 4.5482             & 0.4503             & 25.2371             & 0.1592             & 0.9089             \\
            \textit{w/o} $\mathcal{R}_{\text{refine}}$  & 4.4635             & 0.4338             & 25.3924             & 0.1485             & 0.9119             \\
            \textit{w/o} $\mathcal{R}_{\text{spatial}}$ & \cellsecond 3.7571 & \cellsecond 0.1480 & \cellsecond 26.9385 & \cellsecond 0.1247 & \cellsecond 0.9249 \\
            \textbf{Ours}                               & \cellfirst 3.4671  & \cellfirst 0.1070  & \cellfirst 27.6755  & \cellfirst 0.1077  & \cellfirst 0.9255  \\
            \bottomrule
        \end{tabular}}
    \label{tab:ablation_raydrop}
\end{table}

\section{More Ablation studies}
\label{sec:suppl_ablations}

\subsection{Impact of ray tracing with Gaussian variants}
Tab.~\ref{tab:ablation_gs} and Fig.~\ref{fig:supp_ablation_gs} show the quantitative and qualitative results of the ablation study on ray tracing with Gaussian variants.
We adopt 3D Gaussians~\cite{Kerbl20233dgs} and 2D Gaussians~\cite{Huang2DGS2024} as our Gaussian primitives for ray tracing separately.
As for 3D Gaussians, we construct the corresponding proxy geometry as an icosahedron, proposed by 3DGRT~\cite{3dgrt2024}.
The results demonstrate that the 2D Gaussians have a slight advantage over 3D Gaussians in terms of rendering quality and efficiency,
which means our ray tracer is compatible with various types of Gaussian primitives and other extensions~\cite{Yu2023MipSplatting,ye2024absgs} applied on Gaussian primitives can be easily integrated into our framework.

\begin{figure*}[th]
    \centering
    \vspace{-4mm}
    \includegraphics[width=1.0\linewidth]{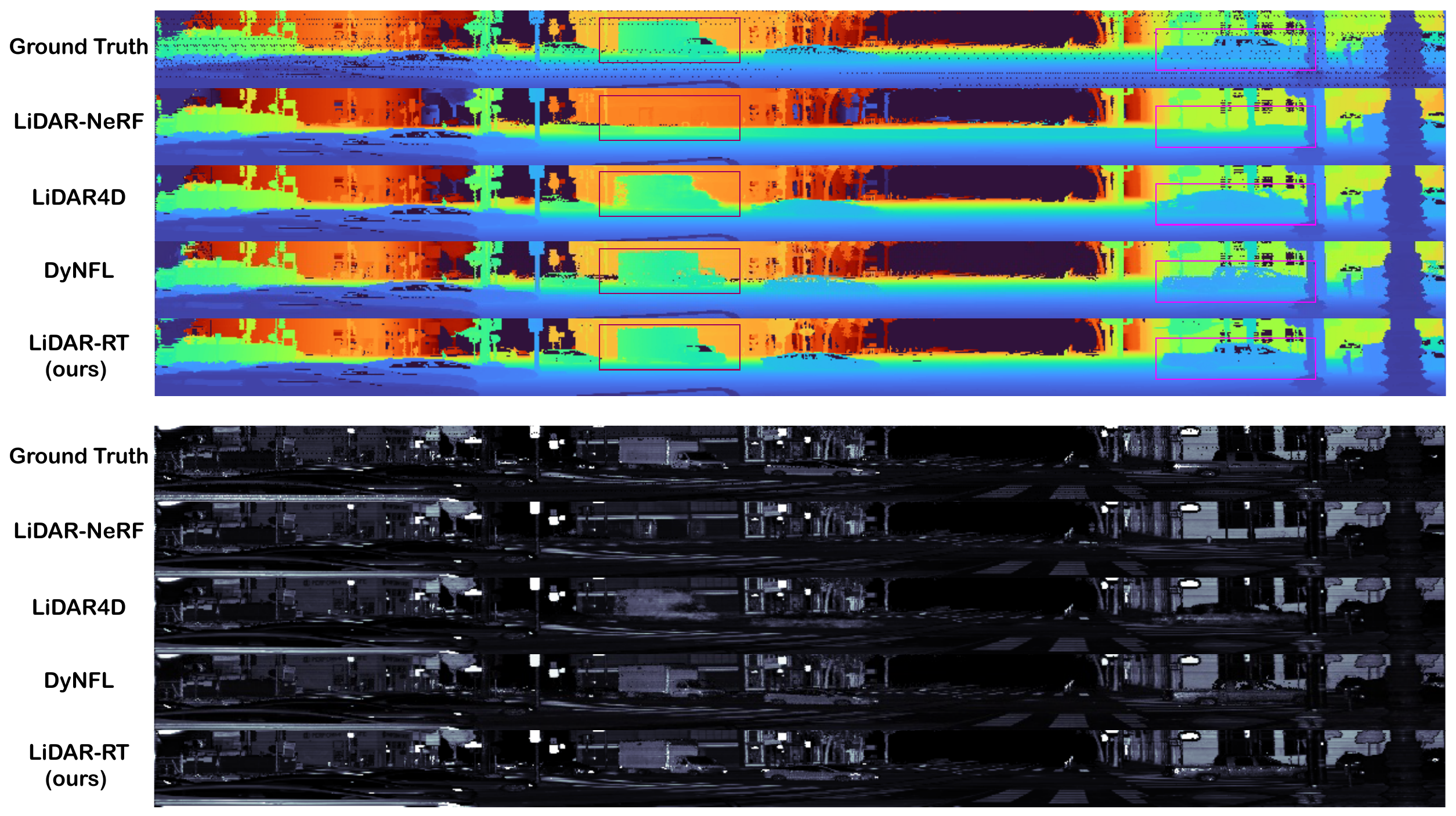}
    \vspace{-4mm}
    \caption{
        \textbf{Qualitative comparison of LiDAR range images on Waymo Open Dataset~\cite{sun2020waymo} sequence seg-132712.}
    }
    \label{fig:supp_waymo1}
    \vspace{-2mm}
\end{figure*}

\begin{figure*}[th]
    \centering
    \includegraphics[width=1.0\linewidth]{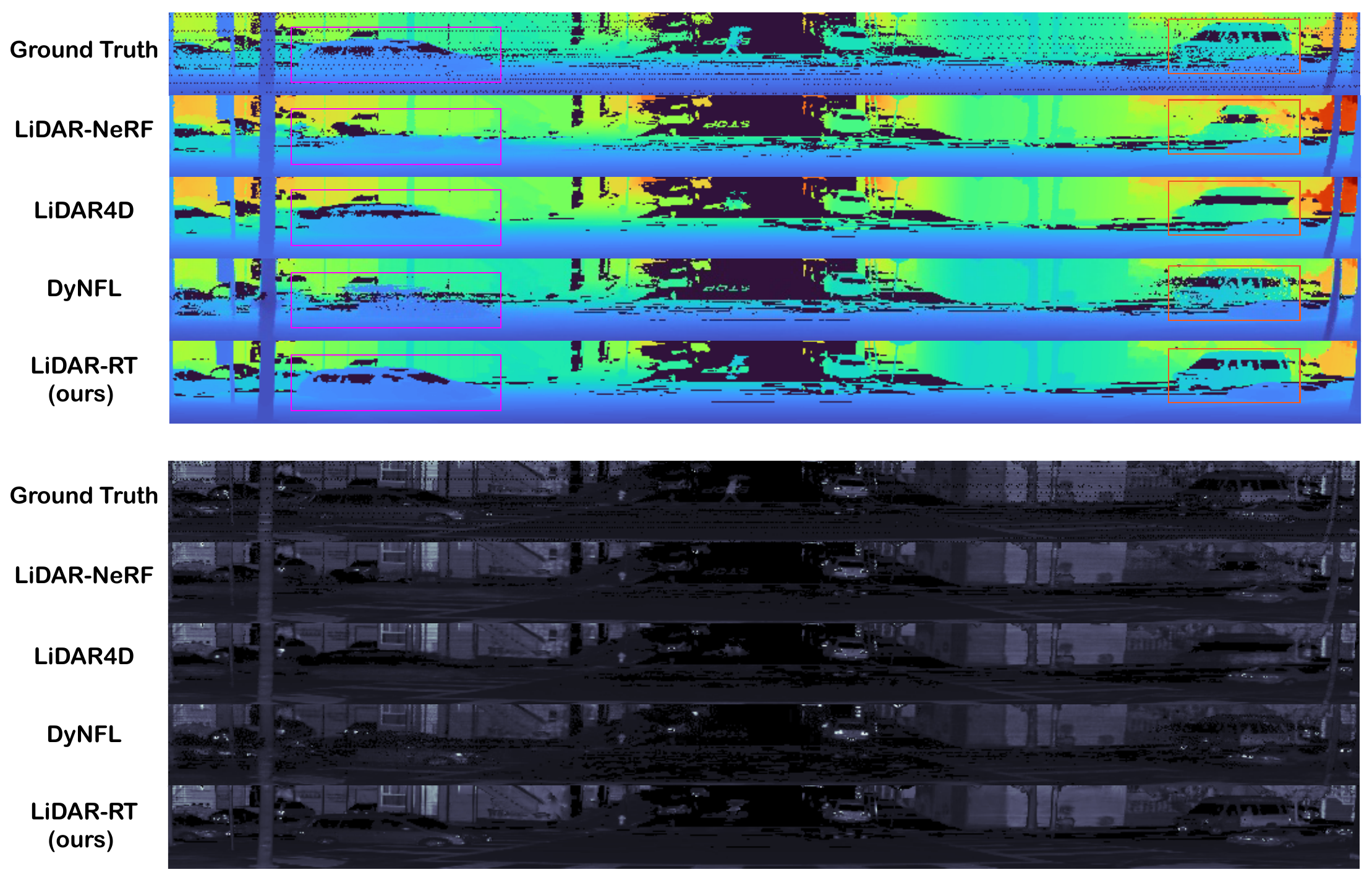}
    \vspace{-6mm}
    \caption{
        \textbf{Qualitative comparison of LiDAR range images on Waymo Open Dataset~\cite{sun2020waymo} sequence seg-108305.}
    }
    \label{fig:supp_waymo2}
    \vspace{-4mm}
\end{figure*}

\subsection{Impact of ray-drop modeling and refinement}

Tab.~\ref{tab:ablation_raydrop} and Fig.~\ref{fig:supp_ablation_raydrop} present the quantitative and qualitative results of our detailed ablation study on ray-drop modeling and refinement.
The variant labeled \textit{w/o} $\mathcal{R}_{\text{hit}}$ models the ray-drop using only a single logit, results in a significant degradation of rendering quality.
The variant \textit{w/o} $\mathcal{R}_{\text{ref}}$ omits the refinement stage,consequently failing to capture the \textit{sensor-level} ray-drop patterns.
Lastly, the \textit{w/o} $\mathcal{R}_{\text{spatial}}$ variant disregards the ray information $(\mathbf{r}_o, \mathbf{r}_d)$ as UNet inputs, leading to a loss of details on dynamic objects.

\section{Additional results}
\label{sec:exp_results}

We provide additional qualitative results on the Waymo dataset~\cite{sun2020waymo} with multiple baselines, as shown in Fig.~\ref{fig:supp_pcd_waymo}, Fig.~\ref{fig:supp_waymo1}, and Fig.~\ref{fig:supp_waymo2}.
The dynamic vehicles are highlighted with colored bounding boxes ($\boldsymbol{\textcolor{pink}{\Box}} / \boldsymbol{\textcolor{orange}{\Box}}$) for better visualization.
Even on the challenging Waymo~\cite{sun2020waymo} dataset with multiple moving actors and the complex urban environment, our \methodnameblank still generates realistic novel LiDAR views with accurate geometry and high-frequency details of dynamic objects.
In contrast, LiDAR-NeRF~\cite{tao2023lidar} struggles with dynamic objects due to its lack of temporal modeling.
LiDAR4D~\cite{zheng2024lidar4d} produces blurry and distorted results on this challenging dataset.
While DyNFL~\cite{Wu2023dynfl} renders plausible results, also exhibits some artifacts around the dynamic objects due to the inaccurate estimations of ray-drop.

\section{Applications}
\label{sec:suppl_applications}

\begin{figure}[th]
    \centering
    \includegraphics[width=1.0\linewidth]{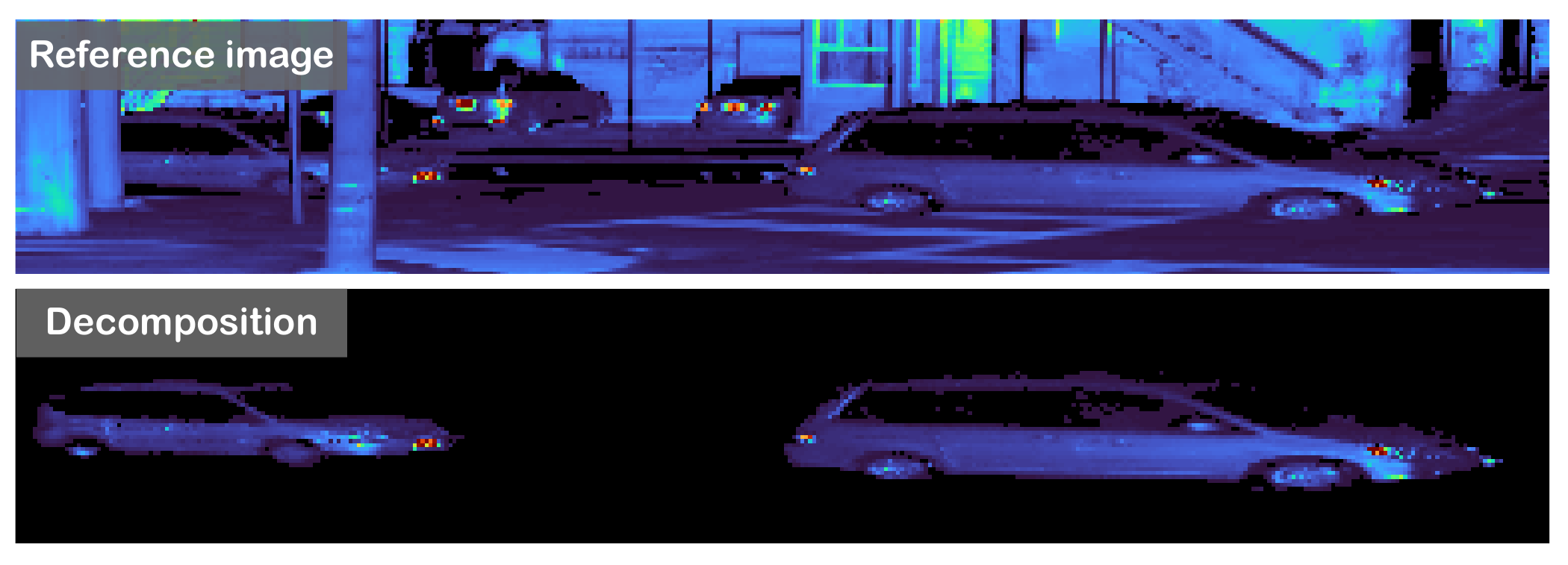}
    \vspace{-6mm}
    \caption{
        \textbf{Decomposition results on Waymo dataset~\cite{sun2020waymo}.} The points are colorized by intensity values from \textcolor{blue}{blue(0)} to \textcolor{red}{red (1)}.
    }
    \vspace{-2mm}
    \label{fig:supp_application_decomposition}
\end{figure}

\PAR{Object decomposition.}
Fig.~\ref{fig:supp_application_decomposition} illustrates the object decomposition results on Waymo dataset~\cite{sun2020waymo}.
Our method is capable of decomposing the foreground dynamic objects clearly and produces high fidelity rendering results.

\end{document}